\documentclass[11pt,a4paper]{article}
\usepackage[hyperref]{style/emnlp2020}
\usepackage{times}
\usepackage{latexsym}
\usepackage{multirow}

\usepackage{microtype}

\aclfinalcopy \def\aclpaperid{585} 
\newif\ifcomment\commentfalse

\usepackage{amsfonts}
\usepackage{amsmath}
\usepackage{amssymb}
\usepackage{bm}
\usepackage{booktabs}
\usepackage[normalem]{ulem}
\newcommand{\abr}[1]{\textsc{#1}}
\usepackage[T1]{fontenc}
\usepackage{graphicx}
\usepackage{multirow}
\usepackage{multicol}

\newcommand{\figfile}[1]{2020_emnlp_clime/figures/#1}
\newcommand{\autofig}[1]{2020_emnlp_clime/auto_fig/#1}

\definecolor{lightblue}{HTML}{3cc7ea}

\newcommand{\name}[0]{\abr{clime}} \newcommand{\vect}[1]{\bm{\mathbf{#1}}}
\newcommand{\norm}[1]{\left\lVert#1\right\rVert}

\ifcomment
\newcommand{\mzcomment}[1]{ \colorbox{lightblue}{   \parbox{.8\linewidth}{ MZ: #1}  }}
\else
\newcommand{\mzcomment}[1]{}
\fi

\usepackage{array}
\newcolumntype{L}[1]{>{\raggedright\let\newline\\\arraybackslash\hspace{0pt}}m{#1}}
\newcolumntype{C}[1]{>{\centering\let\newline\\\arraybackslash\hspace{0pt}}m{#1}}
\newcolumntype{R}[1]{>{\raggedleft\let\newline\\\arraybackslash\hspace{0pt}}m{#1}}

\usepackage{subcaption}
\usepackage[export]{adjustbox}

\title{Interactive Refinement of Cross-Lingual Word Embeddings}

\author{
    Michelle Yuan\thanks{$^{\star}$ indicates equal contribution} \\
    University of Maryland \\
    {\tt myuan@cs.umd.edu} \\\And
    Mozhi Zhang\footnotemark[1] \\
    University of Maryland\\
    {\tt mozhi@cs.umd.edu} \\\And
    Benjamin Van Durme \\
    Johns Hopkins University \\
    {\tt vandurme@jhu.edu} \\\AND
    Leah Findlater \\
    University of Washington\\
    {\tt leahkf@uw.edu} \\\And
    Jordan Boyd-Graber \\
    University of Maryland \\
    {\tt jbg@umiacs.umd.edu} \\
}

\begin{document}

\maketitle
\begin{abstract}
    
Cross-lingual word embeddings transfer knowledge between languages: models trained on high-resource languages can predict in low-resource languages.
We introduce \name{}, an interactive system to
quickly refine cross-lingual word embeddings for a given classification problem.
First, \name{} ranks words by their salience to the downstream task.
Then, users mark similarity between keywords and their nearest neighbors in the embedding space.
Finally, \name{} updates the embeddings using the annotations.
We evaluate \name{} on
identifying health-related text in four low-resource languages: Ilocano, Sinhalese, Tigrinya, and Uyghur.
Embeddings refined by \name{} capture more \emph{nuanced} word semantics and have higher test accuracy than the original embeddings.
\name{} often improves accuracy faster than an active learning baseline and can be easily combined with active learning to improve results.

\end{abstract}

\section{Introduction}
\label{sec:intro}

\begin{figure}
    \centering
    \includegraphics[width=\linewidth]{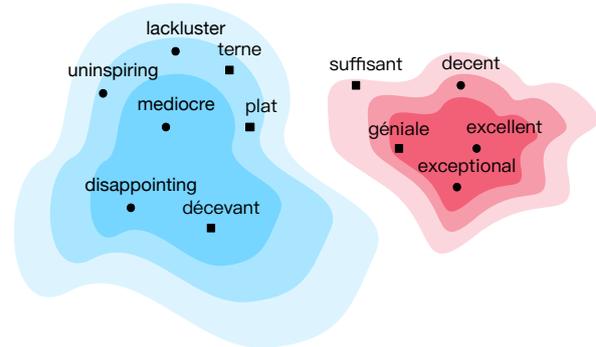}
    \caption{A hypothetical topographic map of an English--French embedding space
    tailored for sentiment analysis. Dots are English words, and squares are
    French words. Positive sentiment words are grouped in a clime (red),     while negative sentiment words     are grouped in another clime (blue). These climes help sentiment analysis.}
    \label{fig:map}
\end{figure}

Modern text classification requires large
labeled datasets and pre-trained word
embeddings~\citep{kim-14,iyyer-15-fixed,joulin-17}.
However, scarcity of both labeled and unlabeled data holds back applications in low-resource languages.
Cross-lingual word embeddings~\citep[\abr{clwe}]{mikolov-13b} can bridge the
gap by mapping words from different languages to a shared vector space.
Using \abr{clwe} features, models trained in a resource-rich language
(e.g., English) can predict labels for other languages.

The success of \abr{clwe} relies on the domain and quality of training
data~\citep{sogaard-18}.
While these methods have impressive word translation accuracy,
they are not tailored for downstream tasks such as text
classification~\citep{glavas-19,zhang-20}.
We develop \textbf{CL}assifying \textbf{I}nteractively with
\textbf{M}ultilingual \textbf{E}mbeddings (\name{}), that efficiently
specializes \abr{clwe} with \emph{human interaction}.\footnote{\url{https://github.com/forest-snow/clime-ui}}~Given a pre-trained \abr{clwe}, a bilingual speaker in the loop reviews the nearest-neighbor words.
\name{} capitalizes on the intuition that neighboring words in an ideal
embedding space should have similar semantic attributes.

In an analogy to geographic \emph{climes}---zones with distinctive
meteorological features---we call areas in the embedding space where words share
similar semantic features climes.
Our goal is to convert neighborhoods in the embedding space into
\emph{classification climes} with words that induce similar labels for a given
classification task.
For example, in the embedding for English--French sentiment
analysis, positive sentiment words such as ``excellent'', ``exceptional'', and
their French translations are together, while ``disappointing'', ``lackluster'', and their translations cluster together elsewhere~(Figure~\ref{fig:map}).
Curating words in the embedding space and refining climes should help downstream classifiers.

First, \name{} uses loss gradients in downstream tasks to find keywords with high salience (Section~\ref{ssec:rank}).
Focusing on these keywords allows the user to most efficiently refine \abr{clwe} by marking their similarity or dissimilarity (Section~\ref{ssec:interaction}).
After collecting annotations, \name{} pulls similar words closer and pushes dissimilar words apart
(Section~\ref{sec:update}), establishing desired climes (Figure~\ref{fig:map}).

Quickly deploying cross-lingual \abr{nlp} systems is particularly important in
global public health emergencies, so we evaluate \name{} on a cross-lingual
document classification task for four low-resource languages: Ilocano,
Sinhalese, Tigrinya, and Uyghur (Section~\ref{sec:experiment}).
\name{} is effective in this low-resource setting because a bilingual speaker can significantly increase test accuracy on identifying health-related documents in
less than an hour.

\name{} is related to active learning~\citep{settles-09}, which also improves
a classifier through user interaction.
Therefore, we compare \name{} with an active learning baseline that asks a
user to label target language documents.
Under the same annotation time constraint, \name{} often has higher accuracy.
Furthermore, the two methods are complementary.
Combining active learning with \name{} increases accuracy even more,
and the user-adapted model is competitive with a large, resource-hungry multilingual transformer~\citep{conneau-20}.

\section{Interactive Neighborhood Reshaping}
\label{sec:ui}

\begin{figure*}
\centering
\includegraphics[width=.8\linewidth]{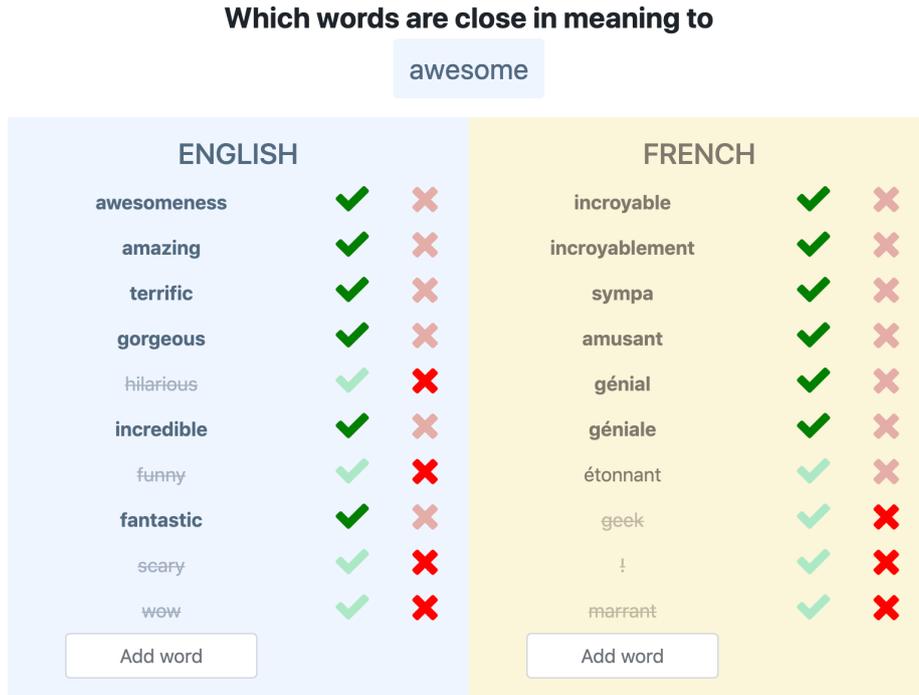}
\caption{\label{fig:ui} The \name{} interface displays a keyword on top while
    its nearest neighbors in the two languages appear in the two columns below.
    A user can accept or reject each neighbor, and add new neighbors by typing
    them in the ``add word'' textboxes.  They may also click on any word to read
    its context in the training set.}
\end{figure*}

This section introduces the interface designed to solicit human feedback on
neighborhoods of \abr{clwe} and our keyword selection criterion.
Suppose that we have two languages with vocabulary
$\mathcal{V}_1$ and~$\mathcal{V}_2$.  Let $\vect{E}$ be a pre-trained \abr{clwe} matrix, where $\vect{E}_w$ is the vector representation of
word type $w$ in the joint vocabulary $\mathcal{V} = \mathcal{V}_1 \cup
\mathcal{V}_2$.
Our goal is to help a bilingual novice (i.e., not a machine learning expert) improve the \abr{clwe} $\vect{E}$ for a downstream task
through inspection of neighboring words.

\subsection{Keyword Selection}
\label{ssec:rank}

With limited annotation time, users cannot vet the entire vocabulary.
Instead, we need to find a small salient subset of \emph{keywords} $\mathcal{K}
\subseteq \mathcal{V}$ whose embeddings, if vetted, would most improve a
downstream task.
For example, if the downstream task is sentiment analysis, our keywords set
should include sentiment words such as ``good'' and ``bad''.
Prior work in active learning solicits keywords using information
gain~\citep{raghavan-06,druck-09,settles-11-fixed}, but this cannot be applied
to continuous embeddings.
\citet{li-16} suggest that the contribution of one dimension of a word
embedding to the loss function can be approximated by the absolute value of its
partial derivative, and therefore they use partial derivatives to visualize the
behavior of neural models.
However, rather than understanding the importance of individual dimensions, we
want to compute the salience of an \emph{entire word vector}.
Therefore, we extend their idea by defining the salience of a word embedding as
the \emph{magnitude} of the loss function's gradient.
This score summarizes salience of all dimensions from a word embedding.
Formally, let $\vect{x} = \langle x_1, x_2, \cdots,
x_n \rangle$ be a document of $n$ words with label $y$; let $L$ be the training
loss function of the downstream model.
We measure the example-level salience of word $x_i$ in document
$\vect{x}$ as
\begin{equation}   S_{\vect{x}}(x_i) = \norm{\nabla_{\vect{E}_{x_i}} L(\vect{x}, y)}_2.
\label{eq:salience1}
\end{equation}

Equation~\ref{eq:salience1} measures the local contribution of a token in one
document, but we are interested in the global importance of a word type across
many documents.
To compute the global salience score of a word type~$w$, we add example-level salience
scores of all token occurrences of a word type $w$ in a large labeled dataset $\vect{X}$ and
multiply by the inverse document frequency~(\abr{idf}) of $w$:
\begin{equation}
S(w) = \abr{idf}(w, \vect{X}) \cdot \sum_{\vect{x} \in
\vect{X}: w \in \vect{x}} S_{\vect{x}}(w).
\label{eq:salience2}
\end{equation}
The \abr{idf} term is necessary because it discounts \emph{stop words} with
high document frequency (e.g., ``the'' and ``of'').
These words are often irrelevant to the downstream task and thus have low example-level salience, but they have high total salience because they appear in many examples.

Based on Equation~\ref{eq:salience2}, we choose the top-$s$ most salient words as the keyword set $\mathcal{K}$.
The hyperparameter $s$ is the number of keywords displayed to the user, which controls the
length of a \name{} session.  We limit $s$ to fifty in experiments.

\subsection{User Interaction}
\label{ssec:interaction}
For each keyword $k$, we want to collect a positive set $\mathcal{P}_k$ with
semantically similar words, and a negative set $\mathcal{N}_k$ with unrelated
words.
To specialize embeddings for a classification task, we ask the user to
consider semantic similarity as \emph{inducing a similar label}.
For example, if the task is English--French sentiment analysis, then ``good'' should be considered similar to ``excellent'' and ``g\'enial''
but dissimilar to ``bad'' and ``d\'ecevant''.
On the interface, the keyword $k$ is displayed on the top, and its nearest neighbors in the
two languages are arranged in two columns (Figure~\ref{fig:ui}).
The neighbors are the words~$w$ with embeddings $\vect{E}_w$ closest to
$\vect{E}_k$ in cosine similarity.
The number of displayed nearest neighbors can be adjusted
as a hyperparameter, which also controls the session length.
For each nearest neighbor, the user can either: (1) press on the green
checkmark to add a positive neighbor to $\mathcal{P}_k$, (2) press on the red
``X'' mark to add a negative neighbor to $\mathcal{N}_k$, or (3) leave an
uncertain neighbor alone.
The ``add word'' textbox lets the user add words that are not in the
current neighbor list.
The added word can then be marked as positive or negative.
Section~\ref{sec:update} explains how \name{} refines the embeddings with the feedback sets
$\mathcal{P}$ and $\mathcal{N}$.
The interface also provides a word concordance---a brief overview of the contexts
where a word appears---to disambiguate and clarify words.
Users can click on any word to find example sentences.

\section{Fitting Word Embeddings to Feedback}
\label{sec:update}

After receiving user annotations, \name{} updates the
embeddings to reflect their feedback.
The algorithm reshapes the neighborhood so that words near a keyword share similar semantic attributes.
Together, these embeddings form desired task-specific
connections between words across languages.  Our update equations are inspired by \abr{attract-repel}~\citep{mrksic-17}, which fine-tunes word embeddings with
synonym and antonym constraints.
The objective in \abr{attract-repel} pulls synonyms
closer to and pushes antonyms further away from their nearest
neighbors.
This objective is useful for large lexical resources like
BabelNet~\citep{navigli-10} with hundreds of thousands linguistic constraints,
but our pilot experiment suggests that the method is not suitable for smaller
constraint sets.
Since \name{} is designed for low-resource languages, we optimize an
objective that reshapes the neighborhood more drastically than
\abr{attract-repel}.

\subsection{Feedback Cost}

For each keyword $k\in\mathcal{K}$, we collect a positive set $\mathcal{P}_k$
and a negative set $\mathcal{N}_k$ (Section~\ref{ssec:interaction}).
To refine embeddings $\vect{E}$ with human feedback, we increase the similarity
between $k$ and each positive word $p\in\mathcal{P}_k$, and decrease the similarity between $k$ and
each negative word $n\in\mathcal{N}_k$.
Formally, we update the embeddings $\vect{E}$ to minimize the following:
\begin{equation}
    C_f(\vect{E}) =
	\sum_{k \in \mathcal{K}} \left(\sum_{n \in \mathcal{N}_k} \vect{E}_k^\top \vect{E}_n - \sum_{p \in \mathcal{P}_k} \vect{E}_k^\top \vect{E}_p\right),
    \label{eq:feedback}
\end{equation}
where $\vect{E}_k^\top \vect{E}_n$ measures the similarity between the keyword $k$ and a negative word $n$, and $\vect{E}_k^\top \vect{E}_p$ measures the similarity between the keyword $k$ and a positive word $p$.  Minimizing $C_f$ is equivalent to maximizing similarities of positive pairs while minimizing similarities of negative pairs.

\subsection{Topology-Preserving Regularization}

Prior embedding post-processing methods emphasize regularization to maintain
the topology---or properties that should be preserved under transformations---of
the embedding space~\citep{mrksic-16,mrksic-17,glavas-18}.
If the original \abr{clwe} brings certain translations together, those
translated words should remain close after updating the embeddings.
The topology also encodes important semantic information that should not be
discarded.
Therefore, we also include the following regularization term:
\begin{equation}
  R(\vect{E}) =  \sum_{w \in \mathcal{V}} \norm{\hat{\vect{E}}_w - \vect{E}_w}_2^2. \label{eq:reg}
\end{equation}
Minimizing $R(\vect{E})$ prevents~$\vect{E}$ from drifting too far away from the original embeddings $\hat{\vect{E}}$.

The final cost function combines the feedback cost (Equation~\ref{eq:feedback}) and the regularizer (Equation~\ref{eq:reg}):
\begin{equation}
    C(\vect{E}) = C_f(\vect{E}) + \lambda R(\vect{E}), \label{eq:cost}
\end{equation}
where the hyperparameter $\lambda$ controls the strength of the regularizer.
The updated embeddings enforce constraints from user feedback while
preserving other structures from the original embeddings.
After tuning in a pilot user study, we set $\lambda$ to one.
We use the Adam optimizer~\citep{kingma-15} with default hyperparameters.

\section{Cross-Lingual Classification Experiments}
\label{sec:experiment}

\begin{table}[t]
    \centering
    \begin{tabular}{p{0.1\textwidth}p{0.3\textwidth}}
	\toprule
	\textbf{Ilocano} &
	... Nagtalinaed dagiti pito a balod ti Bureau of Jail Management and
	Penology (BJMP) ditoy ciudad ti Laoag iti isolation room gapo iti tuko ...\\
	\midrule
	\textbf{English} &
	... Seven inmates from the Bureau of Jail Management and Penology (BJMP),
	Laoag City, have been transferred to the isolation room due to chicken
	pox ...\\
	\bottomrule
  \end{tabular}
  \caption{Excerpt of a positive Ilocano test example (top) and its
  English translation (bottom) that describes a medical emergency.}
  \label{tab:example}
\end{table}

\begin{figure*}[t]
    \centering
    \includegraphics[width=0.8\textwidth]{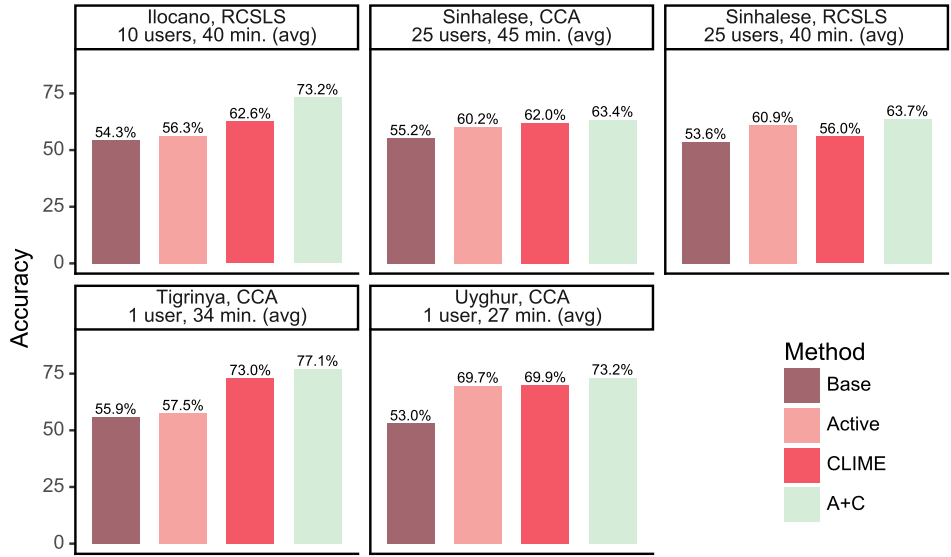}
    \caption{Test accuracy of four methods on four target languages and two
    \abr{clwe} methods.
    \textbf{Base} uses the original \abr{clwe} and the original training
    set.
    \textbf{Active} uses the original \abr{clwe} and a training set
    augmented by active learning.
    We select and label fifty \emph{target language} documents by
    uncertainty sampling and combine them with the source
    language training set.
    \textbf{\name{}} uses the \abr{clwe} refined by \name{} and the
    original training set.
    \textbf{A+C} uses the \abr{clwe} refined by \name{} and a training
    set augmented by active learning.
    We control the number of user interactions so that \textbf{Active}, \textbf{\name{}}, and \textbf{A+C}
    require the similar interaction time~(Section~\ref{ssec:methods}).
    The Sinhalese and Ilocano results are averaged over multiple users, while
    we only have one user for other languages.
    Each subcaption indicates the target language, embedding alignment,
    number of users, and average time per user.
    \textbf{\name{}} has higher accuracy than \textbf{Active} on four of the
    five embeddings, and the combined \textbf{A+C} model has the highest.}
  \label{fig:acc}
\end{figure*}

\begin{figure}[t]
    \centering
	\includegraphics[width=\linewidth]{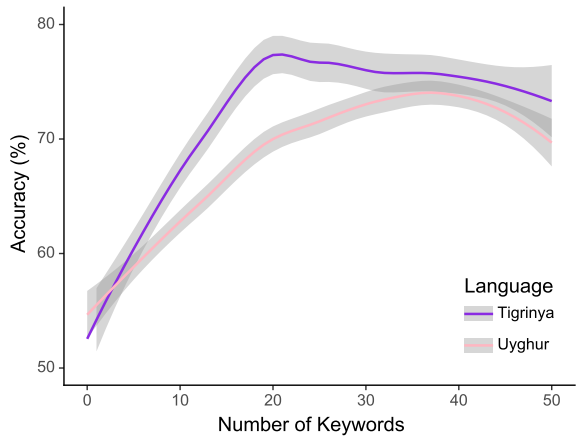}
    \caption{For Uyghur (pink) and Tigrinya (purple), we compare test accuracy between
    sets of \abr{clwe} that differ in the number of keywords used to refine them.
    The leftmost point corresponds to the \textbf{Base} model in Figure~\ref{fig:acc},
    while the rightmost point corresponds to the \textbf{\name{}} model.
    Test accuracy generally improves with more feedback at the beginning but
    slightly drops after reaching an optimal number of keywords.}
    \label{fig:query_acc}
\end{figure}

\begin{figure*}[t]
    \centering
    \begin{subfigure}{\linewidth}
        \centering
        \begin{subfigure}{0.44\linewidth}
            \centering
            \includegraphics[width=\linewidth,frame]{\figfile{plague_nn_1_si.png}}
        \end{subfigure}
        {\LARGE$\xrightarrow{\name}$}
        \begin{subfigure}{0.45\linewidth}
            \centering
            \includegraphics[width=\linewidth,frame]{\figfile{plague_nn_2_si.png}}
        \end{subfigure}
        \caption{Neighborhood of ``plague''}
        \label{fig:plague}
    \end{subfigure}
    \begin{subfigure}{\linewidth}
        \centering
        \begin{subfigure}{0.44\linewidth}
            \centering
            \includegraphics[width=\linewidth,frame]{\figfile{ill_nn_1_si.png}}
        \end{subfigure}
        {\LARGE$\xrightarrow{\name}$}
        \begin{subfigure}{0.45\linewidth}
            \centering
            \includegraphics[width=\linewidth,frame]{\figfile{ill_nn_2_si.png}}
        \end{subfigure}
        \caption{Neighborhood of ``ill''}
        \label{fig:ill}
    \end{subfigure}
    \caption{\abr{t-sne} visualization of embeddings before (left) and after
    (right) \name{} updates.
    From one Sinhalese user study, we inspect two keywords, ``ill'' and
    ``plague'', and their five closest neighbors in English (blue) and
    Sinhalese (green).
    The Sinhalese words are labeled with English translations.
    Shape denotes the type of feedback: ``+'' for positive neighbors and ``x''
    for negative neighbors.}
    \label{fig:nn_analysis}
\end{figure*}

\begin{figure}[t]
    \centering
    \includegraphics[width=0.8\linewidth]{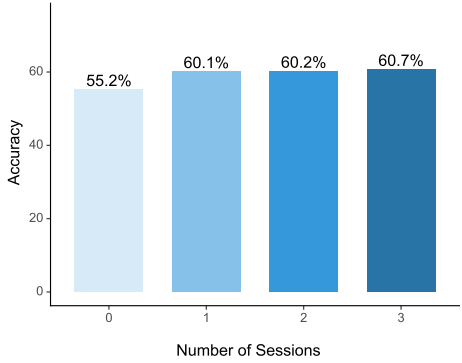}
    \caption{Progress of five Sinhalese users over three \name{} sessions.
    Largest increase in test accuracy occurs after first session.
    The leftmost point is the \textbf{Base} model from Figure~\ref{fig:acc}.
    Average accuracy for first session is not the same as Figure~\ref{fig:acc}
    because only a subset of users are asked to complete three sessions.}
    \label{fig:multi_acc}
\end{figure}

We evaluate \name{} on cross-lingual
document-classification~\citep{klementiev-12}, where we build a text
classifier for a low-resource target language using labeled data in a
high-resource source language through \abr{clwe}.
Our task identifies whether a document describes a medical
emergency, useful for planning disaster relief~\citep{strassel-16}.
The source language is English and the four low-resource
target languages are Ilocano, Sinhalese, Tigrinya, and Uyghur.

Our experiments confirm that a bilingual user can quickly improve the test
accuracy of cross-lingual models through \name{}.
Alternatively, we can ask an annotator to improve the model
by labeling more training documents in the target language.
Therefore, we compare \name{} to an active learning baseline that queries the
user for document labels; \name{} often improves accuracy faster.
Then, we combine \name{} and active learning to show an even faster
improvement of test accuracy.

Comparing active learning to \name{} may seem unfair at first glance.
In theory, document labeling only requires target language knowledge, while
\name{} learns from a bilingual user.
In practice, researchers who speak a high-resource language provide instructions to the annotator and answer their
questions, so bilingual knowledge is usually required in document labeling for
low-resource languages.
Moreover, \name{} is complementary to active learning, as combining them gives
the highest accuracy across languages.

We also experiment with refining the same set of keywords with multiple
rounds of user interaction.
The repeated sessions slightly improve test accuracy on average.
Finally, we compare with \abr{xlm-r}~\citep{conneau-20}, a state-of-the-art
multilingual transformer.
Despite using fewer resources, \name{} has competitive results.

\subsection{Experiment Setup}

\paragraph{Labeled Data.}
We train models on 572 English documents and test on 48 Ilocano documents, 58
Sinhalese documents, 158 Tigrinya documents, and 94 Uyghur documents.
The documents are extracted from \abr{lorelei} language
packs~\citep{strassel-16}, a multilingual collection of documents of emergencies with a public health component.\footnote{Download from \url{https://www.ldc.upenn.edu}}
To simplify the task, we consider a binary classification problem of detecting
whether the documents are
associated with medical needs.
Table~\ref{tab:example} shows an example document.
To balance the label distribution, we sample an equal number of negative
examples.

\paragraph{Word Embeddings.}
To transfer knowledge between languages, we build \abr{clwe} between English
and each target language.
We experiment with two methods to pre-train \abr{clwe}: (1) train monolingual
embeddings with word2vec~\citep{mikolov-13-fixed} and align with
\abr{cca}~\citep{faruqui-15,ammar-16}, (2) train monolingual embeddings with
fastText~\citep{bojanowski-17} and align with \abr{rcsls}~\cite{joulin-18}.
The English embeddings are trained on Wikipedia and the target language embeddings
are trained on unlabeled documents from the \abr{lorelei} language packs.
For alignment, we use the small English dictionary in each pack.
Low-resource language speakers are hard to find, so we do not
try all combinations of languages and \abr{clwe}:
we use \abr{cca} embeddings for Tigrinya and Uyghur, \abr{rcsls}
embeddings for Ilocano.
Since Sinhalese speakers are easier to find, we experiment with both \abr{clwe}
for Sinhalese.

\paragraph{Text Classifier.}
Our classifier is a convolutional neural network~\citep{kim-14}.
Each document is represented as the concatenation of word embeddings and passed
through a convolutional layer, followed by max-pooling and a final softmax
layer.
To preserve cross-lingual alignments, we freeze embeddings during training.
This simple model is effective in low-resource cross-lingual
settings~\citep{chen-18,schwenk-18}.
We minimize cross-entropy on the training set by running Adam~\citep{kingma-15}
with default hyperparameters for thirty epochs.  All experiments use GeForce GTX 1080
\abr{gpu} and 2.6 GHz \abr{amd} Opteron 4180 processor.

\paragraph{User Study.}
We use Upwork to hire participants who are fluent in both English and the
target language.\footnote{\url{https://upwork.com/}}
Low-resource language speakers are hard to find, so we have a different number of
users for each language.
We hire ten Ilocano users and twenty-five
Sinhalese users.
For additional case studies, we hire one Tigrinya user and one Uyghur user.
Each user annotates the fifty most salient keywords, which takes less than an hour (Figure~\ref{fig:acc}).
For each keyword, we show five nearest neighbors for each language.
Each user provides about nine constraints for each keyword.

\subsection{Comparisons}
\label{ssec:methods}

After receiving feedback, we update the embeddings (Section~\ref{sec:update}).
We evaluate the new embeddings by retraining a classifier.
For each set of embeddings, we train ten models with different random seeds and
report average test accuracy.

We compare a classifier trained on the updated embeddings
(\textbf{\name{}} in Figure~\ref{fig:acc}) against two baselines.
The first baseline is a classifier trained on original embeddings
(\textbf{Base} in Figure~\ref{fig:acc}).
If we have access to a bilingual speaker, an alternative to using \name{} is to
annotate more training documents in the target language.
Therefore, we also compare \name{} to uncertainty
sampling~\citep{lewis-1994}, an active learning method that asks a user to
label documents (\textbf{Active} in Figure~\ref{fig:acc}).
We choose a set of fifty documents where model outputs have the
highest entropy from a set of unlabeled \emph{target language} documents
and ask an annotator to label them as additional training documents.
We then retrain a model on both the English training set and the fifty target
language documents, using the original embeddings.
For each model, a human annotator labels fifty
documents within forty to fifty minutes.
This can either be slower or take approximately the same time as an average
\name{} session (Figure~\ref{fig:acc}).
Thus, any improvements in accuracy using \name{} are even more impressive
given that \textbf{Active} is no faster than \name{}.

Finally, we explore combining active learning and \name{}
(\textbf{A+C} in Figure~\ref{fig:acc}).
Document-level and word-level interactions are complementary, so
using both may lead to higher accuracy.
To keep the results comparable, we allocate half of the user interaction time
to active learning, and the other half to \name{}.
Specifically, we use active learning to expand the training set with twenty-five
target language documents and refine the embeddings by running \name{} on
only twenty-five keywords.  Then, we retrain a model using both the augmented
training set and the refined embeddings.

\subsection{Results and Analysis}
\label{ssec:analysis}

\paragraph{Effectiveness of \name{}.}

Figure~\ref{fig:acc} compares the four methods described in the previous
section.
\name{} is effective in this low-resource setting.
On all four target languages, the classifier trained on embeddings refined
by \name{} has higher accuracy than the classifier that trains on the original
embeddings: \name{} reshapes embeddings in a way that helps classification.
\name{} also has higher accuracy than active learning for most users.
The combined method has the highest accuracy: active learning and \name{} are
complementary.
Single-sample $t$-tests confirm that \textbf{\name{}} is significantly
better than \textbf{Base} and \textbf{A+C} is significantly better than
\textbf{Active} (Appendix~\ref{ssec:stats}).

\paragraph{Keyword Detection.}
We inspect the list of the fifty most salient keywords~(Section~\ref{ssec:rank}).
Most keywords have obvious connections to our classification task of detecting
medical emergencies, such as ``ambulance'', ``hospitals'', and ``disease''.
However, the list also contains some words that are unrelated
to a medical emergency, including ``over'' and ``given''.
These words may be biases or artifacts from training data~\citep{feng-18-fixed}.

\paragraph{Number of Keywords.}
To evaluate how feedback quantity changes accuracy, we vary the number of
keywords and compare test accuracy on Tigrinya and Uyghur datasets
(Figure~\ref{fig:query_acc}).
For each keyword~$s$ from one to fifty, we update the original embeddings using
only the feedback on the top-$s$ keywords and evaluate each set of
embeddings with test accuracy.
For both languages, test accuracy generally increases with more annotation at
the beginning of the session.
Interestingly, test accuracy plateaus and slightly drops after reaching an
optimal number of keywords, which is around twenty for Tigrinya and about
forty for Uyghur.
One explanation is that the later keywords are less salient, which causes the
feedback to become less relevant.
These redundant constraints hamper optimization and slightly hurt test accuracy.

\paragraph{Qualitative Analysis.}
To understand how \name{} updates the embeddings, we visualize changes in the
neighborhoods of keywords with t-\abr{sne}~\citep{maaten-08}.
All embeddings from before and after the user updates are projected into the
same space for fair distance comparison.
We inspect the user updates to the Sinhalese \abr{cca}
embeddings~(Figure~\ref{fig:nn_analysis}).
We confirm that positive neighbors are pulled closer and negative neighbors
are pushed further away.
The user marks ``epidemic'' and ``outbreak'' as similar to the keyword
``plague'', and these words are closer after updates~(Figure~\ref{fig:plague}).
For the keyword ``ill'', the user marks ``helpless'' as a negative neighbor,
because ``helpless'' can signal other types of situations and is more ambiguous
for detecting a medical emergency.
After the update, ``helpless'' is pushed away and disappears from the nearest
neighbors of ``ill'' (Figure~\ref{fig:ill}).
However, a few positive neighbors have inadvertently moved away,
such as the Sinhalese translation for ``ill''.
The update algorithm tries to satisfy constraints for multiple keywords, so
soft constraints may be overlooked.
This motivates repeated \name{} sessions where the user can continue
fixing errors.

\subsection{Repeating User Sessions}

We investigate the effects of having a user complete multiple \name{} sessions.
After the user finishes a session, we fit the embeddings to their feedback,
produce a new vocabulary ranking, and update the interface for the next session.
We experiment on the Sinhalese dataset with \abr{cca} embeddings and ask five
users to complete three sessions of fifty keywords.
Average test accuracy increases with more sessions, but the improvement is
marginal after the first session (Figure~\ref{fig:multi_acc}).
By the end of the three sessions, one user reaches 65.2\% accuracy, a
significant improvement from the 55.2\% baseline.

\subsection{Comparing with Contextual Embeddings}
Contextualized embeddings based on multilingual transformers reach
state-of-the-art in many tasks,
so we compare \name{} with these models.
Most existing models~\citep{wu-19,lample-19}
do not cover our low-resource languages.
The only exception is \abr{xlm-r}~\citep{conneau-20}, which covers Uyghur and
Sinhalese.
To compare with \name{}, we fine-tune \abr{xlm-r} for three epochs with AdamW~\citep{loshchilov-2019}, batch size of sixteen, and learning
rate of 2e-5.  We compute average accuracy over ten runs with different random seeds.

For Uyghur, \abr{xlm-r} has lower accuracy than our \textbf{A+C} approach ($71.7\%$ vs. $73.2\%$).
This is impressive given that \abr{xlm-r} uses much more resources:
270 million parameters, 2.5TB of multilingual Common Crawl data, and 500 GPUs.
In contrast, the \textbf{A+C} model has 120K parameters and is built in less
than two hours with a single GPU (including human interaction and model
training).

For Sinhalese, \abr{xlm-r} has higher accuracy than our \textbf{A+C} approach ($69.3\%$ vs. $63.7\%$).
Common Crawl has much more Sinhalese words than Uyghur words.
This aligns with our intuition: \name{} is more useful in low-resource settings, whereas multilingual transformers are more appropriate for languages with more data.
Future work can extend the interactive component of \name{} to multilingual transformers.

\section{Related Work}
\label{sec:related}

\paragraph{Cross-Lingual Word Embeddings.}
\citet{ruder-19} summarize previous \abr{clwe} methods.
These methods learn from \emph{existing} resources such as dictionaries,
parallel text, and monolingual corpora.
Therefore, the availability
and quality of training data primarily determines
the success of these methods~\citep{sogaard-18}. To improve the suitability of \abr{clwe} methods in low-resource settings,
recent work focuses on learning without cross-lingual
supervision~\citep{artetxe-18b,hoshen-18}
and normalizing monolingual embeddings before alignment~\citep{zhang-19}.
In contrast, we design a human-in-the-loop system to efficiently improve
\abr{clwe}.
Moreover, previous \abr{clwe} methods are heavily tuned for the intrinsic
evaluation task of dictionary induction, sometimes to the detriment of
downstream tasks~\citep{glavas-19,zhang-20b}.
Our method is tailored for downstream tasks such as text classification.

\paragraph{Cross-Lingual Document Classification.}
Prior approaches transfer knowledge with cross-lingual resources, such as
bilingual dictionaries~\citep{wu-08,shi-10}, parallel text~\citep{xu-17}, labeled data
from related languages~\citep{zhang-20}, structural
correspondences~\citep{prettenhofer-2010}, multilingual topic
models~\citep{ni-2011,andrade-15}, machine translation~\citep{wan-09-fixed,zhou-16}, and
\abr{clwe}~\citep{klementiev-12}.
Our method instead brings a bilingual speaker in the loop to \emph{actively}
provide cross-lingual knowledge, which is more reliable in low-resource
settings.
Concurrent to our work, \citet{karamanolakis-20} also show that keyword translation is very useful for cross-lingual document classification.

\paragraph{Human-in-the-Loop Multilingual Systems.}
\name{} is inspired by human-in-the-loop systems that bridge language gaps.
\citet{brown-2016} build an interactive translation platform to help refugee
resettlement.
\citet{yuan-18} interactively align topic models across languages.

\paragraph{Active Learning.}
A common solution to data scarcity is active learning, the framework in which
the learner iteratively queries an oracle (often a human) to receive
annotations on unlabeled data.
\citet{settles-09} summarizes popular active learning methods.
Most active learning methods solicit labels for training examples/documents,
while \name{} asks for word-level annotation.
Previous active learning methods that use feature-level
annotation~\citep{raghavan-06,zaidan-07,druck-09,settles-11-fixed} are not applicable to
neural networks and \abr{clwe}.
Closely related to our work, \citet{yuan-2020-alps} propose an active learning strategy that selects examples based on language modeling pre-training.

\paragraph{Neural Network Interpretation.}
Our keyword detection algorithm expands upon prior
work in interpreting neural networks.
\citet{li-16} uses the gradient of the objective function to linearly
approximate salience of one dimension, which helps interpret and visualize
word compositionality in neural networks.
Their ideas are inspired by visual salience in computer vision~\citep{simonyan-2013,zeiler-14}.
We further extend the idea to compute the global salience of an entire word vector
across a labeled dataset.

\paragraph{Specializing Word Embeddings.}
Our update equations modify prior work on specializing word
embeddings that are designed to improve word embeddings with a large
lexical knowledge base.
\citet{faruqui-15} \emph{retrofit} word embeddings to synonym constraints.
\citet{mrksic-16} expand the method by also fitting antonym relations.
\citet{mrksic-17} includes both monolingual and cross-lingual constraints to
improve \abr{clwe}.
\citet{glavas-18} use a neural network to learn an specialization
function that generalize to words with no lexical constraints.
Closest to our work, \citet{zhang-20b} retrofit \abr{clwe} to dictionaries and observe improvement in downstream tasks.

\section{Conclusion and Future Work}

\name{} is an interactive system that enhances \abr{clwe} for a
 task by asking a bilingual speaker for word-level similarity
annotations.
We test \name{} on cross-lingual information triage in international health emergencies for four
low-resource languages.
Bilingual users can quickly improve a model with the help of \name{} at a
faster rate than an active learning baseline.
Combining active learning with \name{} further improves the system.

\name{} has a modular design with three components: keyword ranking, user
interface, and embedding refinement.
The keyword ranking and the embedding refinement modules build upon existing
methods for interpreting neural networks~\citep{li-16} and fine-tuning word
embeddings~\citep{mrksic-17}.
Therefore, future advances in these areas may also improve \name{}.
Another line of future work is to investigate alternative user interfaces.
For example, we could ask bilingual users to \emph{rank} nearest
neighbors~\citep{sakaguchi-18} or provide scalar grades~\citep{hill-15} instead
of accepting/rejecting individual neighbors.

We also explore a simple combination of active learning and
\name{}. Simultaneously applying both methods is better than using either alone.
In the future, we plan to train a policy that dynamically combines the two interactions with reinforcement learning~\citep{fang-17}.

\section*{Acknowledgments}

We appreciate the feedback from Joe Barrow, Shi Feng, Chen Zhao, and the anonymous reviewers.
We thank Pedro Rodriguez, Jo Shoemaker, and Craig Harman for helping with the user study.
Michelle Yuan is supported by JHU Human Language Technology Center of Excellence (HLTCOE).
This research is supported in part by the Office of the Director of National Intelligence (ODNI), Intelligence Advanced Research Projects Activity (IARPA), via the BETTER Program contract \#2019-19051600005. The views and conclusions contained herein are those of the authors and should not be interpreted as necessarily representing the official policies, either expressed or implied, of ODNI, IARPA, or the U.S. Government. The U.S. Government is authorized to reproduce and distribute reprints for governmental purposes notwithstanding any copyright annotation therein.

\bibliography{bib/journal-full,bib/jbg,bib/michelle,bib/mozhi}
\bibliographystyle{style/acl_natbib}

\clearpage

\appendix
\section{Appendices}

\subsection{Statistical Significance}
\label{ssec:stats}

\begin{table}[h]
    \centering
    \begin{tabular}{llrrr}
    \toprule
    Comparison & Model  & $p$ & $t$ & $df$ \\
    \midrule
    \multirow[t]{3}{4em}{\name{} vs.  \textbf{Base}}  &  \abr{si(cca)} & <0.01  & 7.64 & 24 \\
                                                & \abr{si(rcsls)} &
        <0.01 & 3.62  & 24  \\
                                                & \abr{il(rcsls)} & <0.01 & 5.16
                                                & 9 \\ \midrule
        \multirow[t]{3}{4em}{\name{} vs. \textbf{Active}}  & \abr{si(cca)} &
        0.07 & 2.00 & 24 \\
                                                & \abr{si(rcsls)} & <0.01 & -7.09 & 24 \\
                                                & \abr{il(rcsls)} &
       <0.01 & 3.96 & 9 \\ \midrule
        \multirow[t]{3}{4em}{\textbf{A+C} vs. \textbf{Active}}  & \abr{si(cca)}
                                                          & <0.01  & 4.297 & 24 \\
                                                & \abr{si(rcsls)} &
        <0.01 & 3.40 & 24 \\
                                                & \abr{il(rcsls)} &
        <0.01 & 13.97 & 9 \\
    \bottomrule
    \end{tabular}
    \caption{Results of single-sample $t$-tests between \name{} and
    \textbf{Base}, \name{} and \textbf{Active}, and \textbf{A+C} and
    \textbf{Active}, showing the $p$-value, the $t$ statistic, and the
    degree of freedoms $df$.
    \name{} is significantly better
    than \textbf{Base}, and \textbf{A+C} is significantly
    better than \textbf{Active} across different languages and embedding models.
    The only combination with results that are not significantly
    different is \name{} and \textbf{Active} for Sinhalese
    (\abr{cca}).
    }
    \label{tab:t}
\end{table}

We run single-sample $t$-tests with $.05$ significance level to see whether
adding word-level annotations with $\name{}$ can significally improve
classification accuracy.
We compare \name{} against \textbf{Base}, \name{} against \textbf{Active}, and
\textbf{A+C} against \textbf{Active}.  We use the user study results from the
Sinhalese models (both \abr{cca} and \abr{rcsls}) and the Ilocano model.
Table~\ref{tab:t} shows that
\name{} is not
significantly different from \textbf{Active} for the Sinhalese \abr{cca}
embeddings but does significantly improve accuracy for the Ilocano model.
Overall, \name{} is significantly
different from \textbf{Base} and \textbf{A+C} is significiantly different from
\textbf{Active} across the experiments for all models.

\end{document}